\documentclass[letterpaper]{article} % DO NOT CHANGE THIS
\usepackage{aaai24}  % DO NOT CHANGE THIS
\usepackage{times}  % DO NOT CHANGE THIS
\usepackage{helvet}  % DO NOT CHANGE THIS
\usepackage{courier}  % DO NOT CHANGE THIS
\usepackage[hyphens]{url}  % DO NOT CHANGE THIS
\usepackage{graphicx} % DO NOT CHANGE THIS
\usepackage{natbib}  % DO NOT CHANGE THIS AND DO NOT ADD ANY OPTIONS TO IT
\usepackage{caption} % DO NOT CHANGE THIS AND DO NOT ADD ANY OPTIONS TO IT
\usepackage{algorithm}
\usepackage{algorithmic}

\usepackage{newfloat}
\usepackage{listings}
\DeclareCaptionStyle{ruled}{labelfont=normalfont,labelsep=colon,strut=off} % DO NOT CHANGE THIS
\floatstyle{ruled}
\newfloat{listing}{tb}{lst}{}
\floatname{listing}{Listing}
\usepackage{multirow}

\title{Task Aware Modulation using Representation Learning for Upsaling of Terrestrial Carbon Fluxes}
\author{
    Aleksei Rozanov\textsuperscript{\rm 1}\equalcontrib,
    Arvind Renganathan\textsuperscript{\rm 1}\equalcontrib,
    Vipin Kumar\textsuperscript{\rm 1} 
}
\affiliations{
    \textsuperscript{\rm 1}University of Minnesota, Twin Cities\\
    \{rozan012, renga016, kumar001\}@umn.edu
}

\begin{document}

\maketitle

\begin{abstract}
Accurately upscaling terrestrial carbon fluxes is central to estimating the global carbon budget, yet remains challenging due to the sparse and regionally biased distribution of ground measurements. Existing data-driven upscaling products often fail to generalize beyond observed domains, leading to systematic regional biases and high predictive uncertainty. We introduce Task-Aware Modulation with Representation Learning (TAM-RL), a framework that couples spatio-temporal representation learning with knowledge-guided encoder-decoder architecture and loss function derived from the carbon balance equation. Across 150+ flux tower sites representing diverse biomes and climate regimes, TAM-RL improves predictive performance relative to existing state-of-the-art datasets, reducing RMSE by 8–9.6\% and increasing explained variance ($R^2$) from 19.4\% to 43.8\%, depending on the target flux. These results demonstrate that integrating physically grounded constraints with adaptive representation learning can substantially enhance the robustness and transferability of global carbon flux estimates.
\end{abstract}

\section{Introduction}

\begin{figure}
    \centering
    \includegraphics[width=0.9\linewidth]{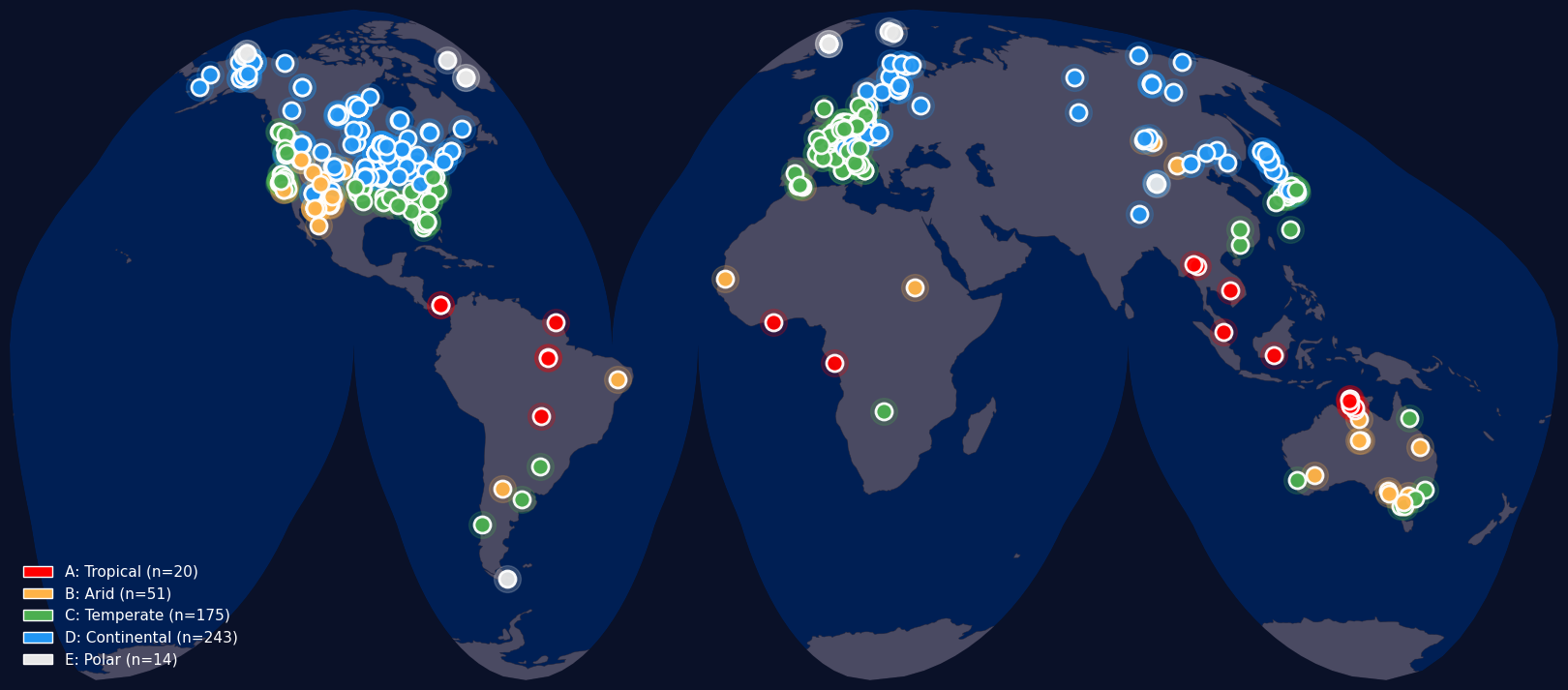}
    \caption{ \small Spatial distribution of the sites used in the current study where color represents climate type according to Köppen–Geiger climate classification \cite{essd-13-5087-2021}. The sites are derived from the FLUXCOM, AmeriFlux, ICOS, and JapanFlux networks.}
    \label{fig:map}
\end{figure}

Accurately quantifying the exchange of carbon dioxide between land and atmosphere is essential for understanding the global carbon balance and predicting climate feedbacks \cite{masson2021summary}. Eddy covariance (EC) towers provide direct \textit{in-situ} measurements of terrestrial carbon fluxes such as net ecosystem exchange (NEE), gross primary production (GPP), and ecosystem respiration (RECO). However, each tower observes only a small area (typically a few square kilometers), making global coverage infeasible. To address this limitation, large-scale networks such as FLUXNET \cite{baldocchi2001fluxnet} and AmeriFlux \cite{novick2018ameriflux} aggregate and distribute EC data from several hundred sites worldwide, with most located in North America and Europe (Fig.~\ref{fig:map}). While EC measurements provide high-fidelity flux estimates at individual sites, their coverage is insufficient to produce spatially continuous assessments of global carbon sequestration needed for climate policy, carbon accounting, and parameterization of Earth system models.

Yet, given this data, an upscaling task can be formulated as infering a continuous flux field from ground-level data and gridded datasets using a non-linear estimator. Classical methods such as resampling and spatial interpolation (bilinear, spline, kriging) perform poorly because they ignore ecosystem context and temporal variability. In contrast, recent advances in representation learning have opened new opportunities for integrating satellite-based observations, which provide dense spatial coverage of vegetation and surface properties, and meteorological data to predict carbon fluxes at regional and global scales. Combining these complementary modalities enables data-driven models to infer fluxes in unobserved regions. Yet, despite impressive results, most ML-based upscaling frameworks remain domain-specific -- trained and evaluated on limited sets of flux towers -- and their generalization to unseen biomes and climates remains weak.

A key reason is that flux–feature relationships are highly context-dependent: the same satellite-derived vegetation index may indicate different physiological states under varying weather conditions. Furthermore, differences in data resolution, noise characteristics, and temporal coverage introduce covariate shifts that standard ML models cannot handle. As a result, current approaches often excel within known regions but fail when transferred globally.

To overcome these limitations, recent research emphasizes domain generalization and task-aware learning approaches that explicitly separate location-specific variability from generalizable ecosystem dynamics. By learning invariant representations that capture process-level dependencies rather than purely statistical correlations, such methods aim to produce models that scale robustly across ecosystems, climates, and temporal contexts.

Thus, here we propose applying the Task-Aware Modulation with Representation Learning (TAM-RL) framework \cite{renganathan2025task} to the problem of upscaling of terrestrial carbon fluxes. The original TAM-RL approach combines an LSTM-based temporal encoder that captures location-specific attributes and produces a latent representation with a modulated decoder that generates the final predictions. Originally developed for few-shot learning, this framework has already demonstrated strong performance across several environmental modeling tasks.
Our work makes the following novel contributions:
\begin{itemize}
\item We extend TAM-RL to the domain generalization setting, evaluating its ability to perform global carbon flux upscaling without site-specific fine-tuning, e.g. zero-shot;
\item We design a knowledge-guided loss function incorporating the carbon balance equation ($NEE = GPP - RECO$), quality-control weighting, and class balancing across ecosystem types;
\item We demonstrate significant performance gains over current state-of-the-art analog, achieving an average RMSE reduction of 8–9.6\% and an $R^2$ increase of 19.4–43.8\% across all ecosystem types.
\end{itemize}
Overall, our results show that TAM-RL effectively addresses the domain generalization challenge and outperforms classical tree-based models that depend on extensive manual feature engineering.

\section{Related Works}
\subsection{Knowledge-Guided ML}
% Knowledge-Guided ML: overview of prior works combining ML with domain constraints (physics-informed NN, hybrid modeling, etc.).

Despite the widespread use of process-based models in Earth system science, their performance is often limited by structural biases, incomplete process representations, and high computational demands. As an alternative, data-driven machine learning (ML) approaches have gained substantial traction over the past decade, offering high efficiency during inference, flexibility in incorporating diverse input drivers, and the ability to learn rich feature representations -- particularly when neural networks (NNs) are employed. However, purely data-driven methods tend to overfit and exhibit poor generalization in data-sparse setups. Knowledge-Guided Machine Learning (KGML) has emerged as a promising paradigm that combines the strengths of process-based and data-driven modeling, enabling faster convergence, improved physical consistency, and better generalization.

According to \cite{karpatne2024knowledge}, strategies for incorporating scientific knowledge into ML can be grouped into three main categories:
\begin{enumerate}
    \item[(a)] Modifying the learning algorithm, for instance by embedding domain constraints directly into the loss function;
    \item[(b)] Integrating scientific knowledge into the model architecture, such as by constraining intermediate layers or designing physics-informed structures;
    \item[(c)] Knowledge-guided weight initialization, for example through pre-training on outputs of physical or “surrogate” models.
\end{enumerate}

Due to space constraints, we forgo a comprehensive review of related work. KGML has been widely explored across scientific domains, including chemistry \cite{ji2021autonomous, wu2023application}, physics \cite{cai2021physics, ren2022phycrnet}, hydrology \cite{read2019process, xu2015data}, and climate science \cite{liu2024knowledge}, demonstrating its broad applicability for learning system dynamics, surrogate modeling, and prior-informed frameworks.

\subsection{Carbon Flux Upscaling}
Carbon flux upscaling methods typically derive targets from EC networks and predict GPP, RECO, and/or NEE from remote sensing and meteorological inputs. The most prominent framework is FLUXCOM \cite{jung2020scaling} and its successor FLUXCOM-X-BASE \cite{nelson2024x}, which uses XGBoost \cite{chen2016xgboost} with MODIS and climate forcings at hourly, 0.05° resolution — we use it as our state-of-the-art baseline. Recent work has explored meta-learning \cite{nathaniel2023metaflux}, PFT-specialized mixture of experts \cite{yuan2025global}, and KGML approaches \cite{fan2025estimating}, but to our knowledge none has framed $\mathrm{CO_2}$ upscaling as a zero-shot regression transfer learning problem.

\begin{figure}[h]
    \centering
    \includegraphics[width=1\linewidth]{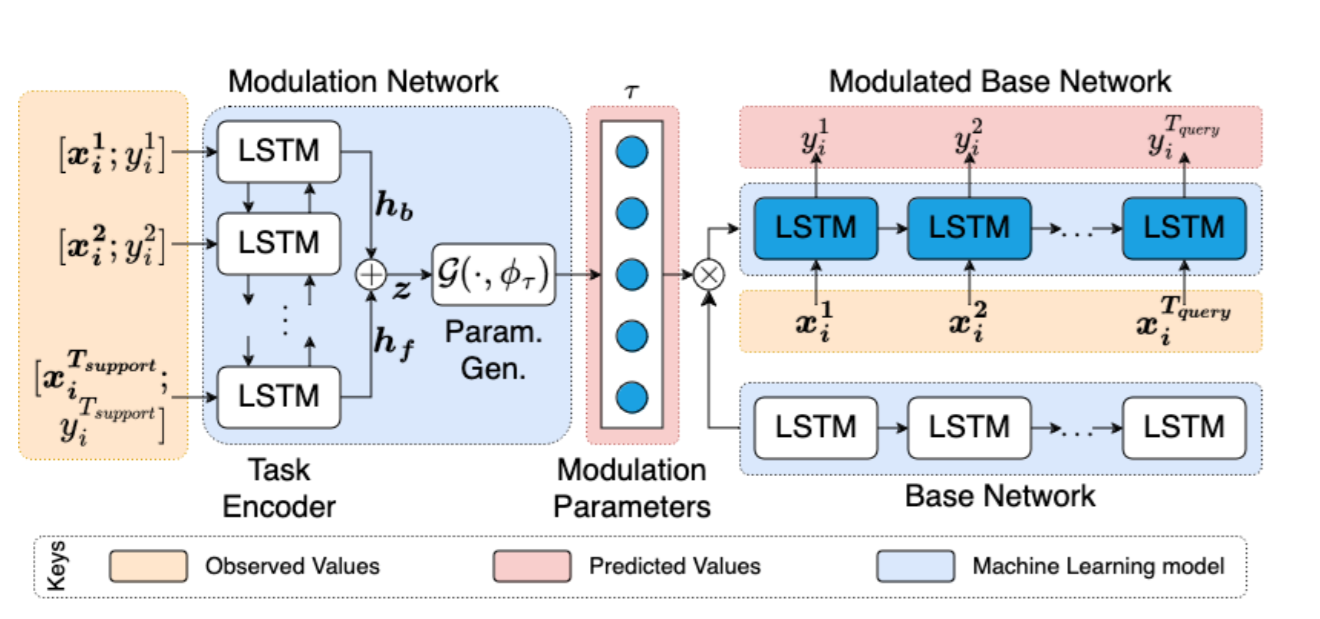}
    \caption{\small TAM-RL Architecture}
    \vspace{-0.5cm}
    \label{fig:tamrl}
\end{figure}

\section{Methodology}

\begin{figure*}[t]
    \centering
    \includegraphics[width=0.9\linewidth]{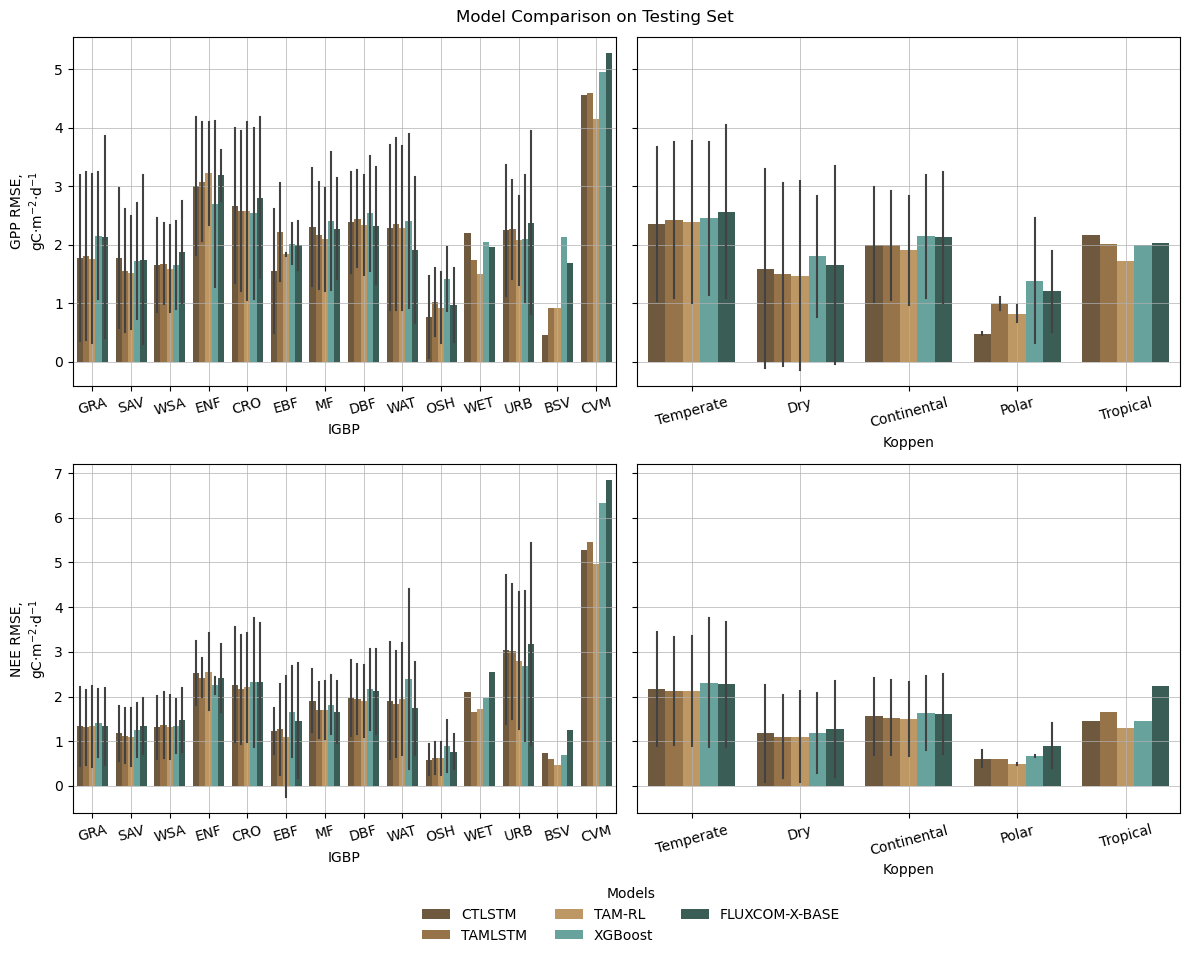}
    \caption{\small Barplots illustrating mean RMSE of CT-LSTM, TALMSTL, TAM-RL, XGBoost and FLUXCOM-X-BASE aggregated by IGBP and Köppen climate types. The error bars are computed for classes with more than one site presenting them. NEE\_VUT\_USTAR50\_QC=1.}
    \vspace{-0.2cm}
    \label{fig:bars}
\end{figure*}

\subsection{Dataset}
Our dataset integrates eddy covariance (EC) fluxes, MODIS satellite observations, and ERA5-Land reanalysis data. Ground-truth fluxes are derived from 579 EC sites (2000--2023) across FLUXNET, AmeriFlux, ICOS \cite{rebmann2018icos}, and JapanFlux \cite{ueyama2025japanflux2024}, spanning all K\"{o}ppen--Geiger climate classes (Fig.~\ref{fig:map}), following the ONEFlux standard \cite{pastorello2020fluxnet2015} for daily GPP\_NT\_VUT\_USTAR50 (GPP) and NEE\_VUT\_USTAR50 (NEE) estimates with NEE\_VUT\_USTAR50\_QC used as a conitnuous quality flag ranging from 0 to 1. Vegetation properties come from MODIS, particularly MOD09GA \cite{mod09ga} and MCD12Q1 \cite{mcd12q1}, at 500~m resolution, extracted over a 2~km~$\times$~2~km tower window via Google Earth Engine \cite{gorelick2017google}. Meteorological predictors are from ERA5-Land \cite{munoz2021era5} at 0.1° resolution. All inputs are harmonized to daily resolution with a 45-day sequence window and 15-day stride. 

\subsection{Loss Function}
We use a composite loss enforcing data quality, class balance, and physical consistency:
\[
\mathcal{L} = \mathrm{MSE} \cdot w_{\mathrm{qc}} \cdot w_{\mathrm{igbp}} \cdot w_{\mathrm{koppen}} + \alpha \cdot L_{\mathrm{flux}}, \quad \alpha=0.1
\]
where $w_{\mathrm{qc}}$ weights samples by their continuous NEE quality flag, $w_{\mathrm{igbp}}$ and $w_{\mathrm{koppen}}$ are inverse-frequency class weights countering spatial imbalance across ecosystem and climate types, and $L_{\mathrm{flux}}$ penalizes violations of $\mathrm{NEE} = \mathrm{RECO} - \mathrm{GPP}$. GPP and RECO are clipped to non-negative values at inference.

\begin{figure*}[t]
    \centering
    \includegraphics[width=0.9\linewidth]{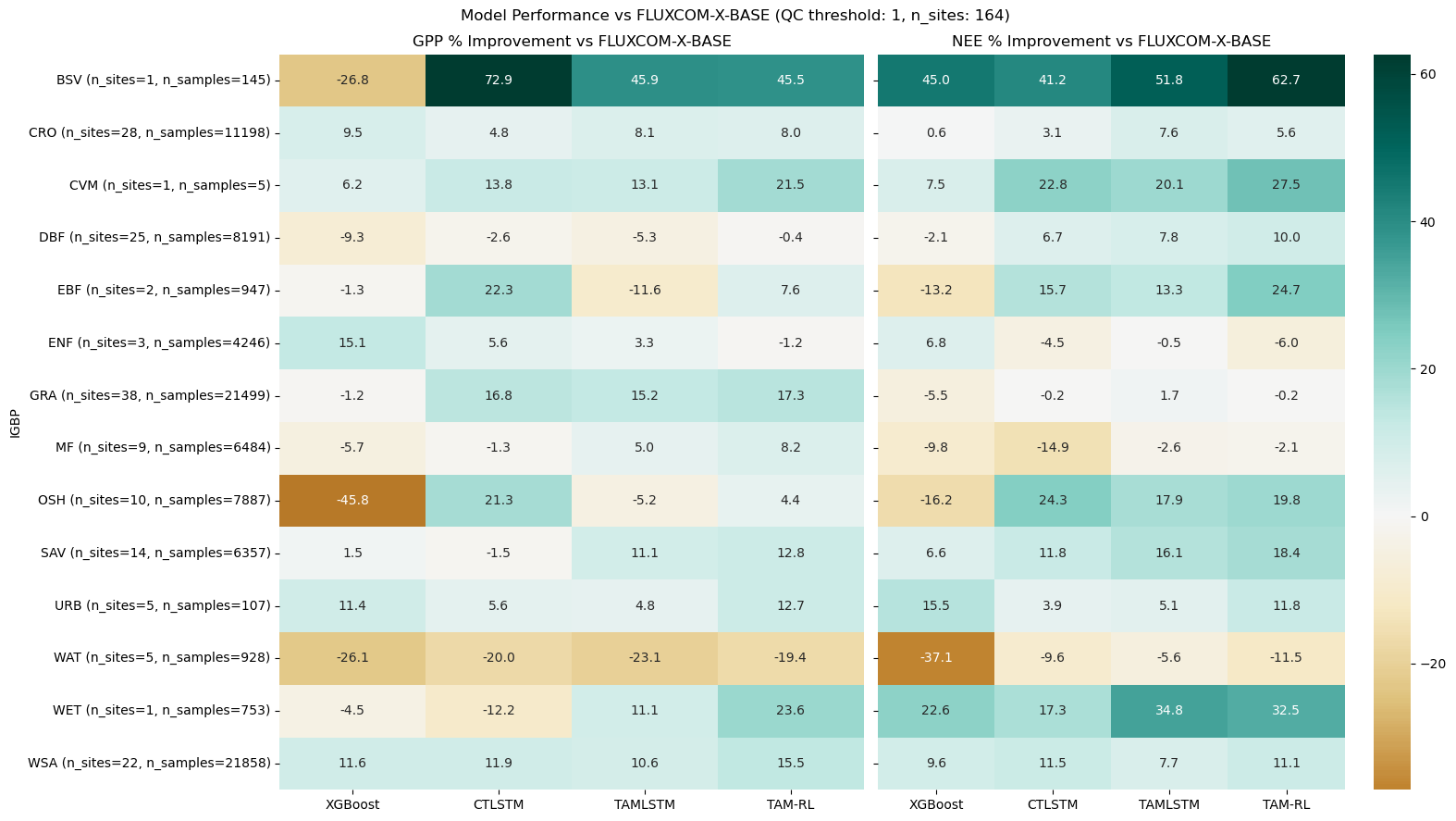}
    \caption{\small Relative RMSE heatmap of XGBoost, CT-LSTM, TAMLSTM, and TAM-RL derived in comparsion to FLUXCOM-X-BASE with NEE\_VUT\_USTAR50\_QC=1.}
    \label{fig:heat}
\end{figure*}

\subsection{Task-Aware Modulation using Representation Learning}
We leverage the TAM-RL framework (Fig.~\ref{fig:tamrl}). The architecture consists of two main components:
\begin{itemize}
    \item \textbf{Modulation network:} Composed of a BiLSTM-based task encoder $\mathcal{E}$ and a modulation parameter generator $G$. The encoder captures temporal information and learns task-specific embeddings $z_{i}$ representing the characteristics of each entity (site). These embeddings are then used by the MLP-based generator $G$ to produce modulation parameters.
    \item \textbf{Forward model $\mathcal{F}$:} A standard LSTM decoder used for conditional generation of sequential responses given the driver data. During inference, the modulation network provides task-specific meta-initialization for $F$.
\end{itemize}

Modulation is a crucial component of TAM-RL, and is applied at two points: the input layer and the final hidden state, using Feature-wise Linear Modulation (FiLM):

$x^{\prime} = \gamma_{1}\bigodot x + \beta_{1}$,

$h^{\prime} = \gamma_{2}\bigodot h + \beta_{2}$,

where $x^{\prime}$ and $h^{\prime}$ is input and hidden state after modulation, $x$ and $h$ input and hidden state before modulation, $\bigodot$ denotes element-wise multiplication, and $\gamma_{i}, \beta_{i}$ are the learned modulation parameters.

Training occurs in two stages:
\begin{enumerate}
    \item \textbf{Pre-training:} Only the decoder network is trained without using any task-specific information. By initializing the model with these pre-trained weights, the decoder is provided with a robust foundation that has shared knowledge across the distribution for subsequent task-specific adaptation.
    \item \textbf{Joint training:} For each task (carbon site), a small support set is used to compute the task embedding. The modulation network converts this embedding into parameters that adjust both the shared feature extractor and prediction head. A separate query set is passed to the adapted decoder, and gradients are backpropagated through both networks. This enables TAM-RL to learn how to modulate representations for new tasks using only a few examples, without test-time fine-tuning.
\end{enumerate}

During inference, historical data for each site (e.g., 2001, 2011, 2021) serves as the support set to derive site context. The modulated base network then generates final predictions for each task.

\section{Results}
We evaluate our models against FLUXCOM-X-BASE as the state-of-the-art product, alongside three additional baselines:
\begin{itemize}
    \item \textbf{CT-LSTM}: a standard LSTM augmented with one-hot encoded static features;
    \item \textbf{TAMLSTM}: the decoder-only component of TAM-RL;
    \item \textbf{XGBoost}: trained on flattened (non-temporal) feature vectors, configured with the same hyperparameters as FLUXCOM-X-BASE for methodological consistency.
\end{itemize}
Each neural network model is represented as a mean ensemble over ten independently trained runs (different random seeds), ensuring comparability with XGBoost, which is ensemble-based by design. FLUXCOM-X-BASE provides hourly GPP and NEE estimates, which we averaged to daily values to match our models' output resolution, along with spatial resolution and grid alignment.

\begin{figure*}[t]
    \centering
    \includegraphics[width=1\linewidth]{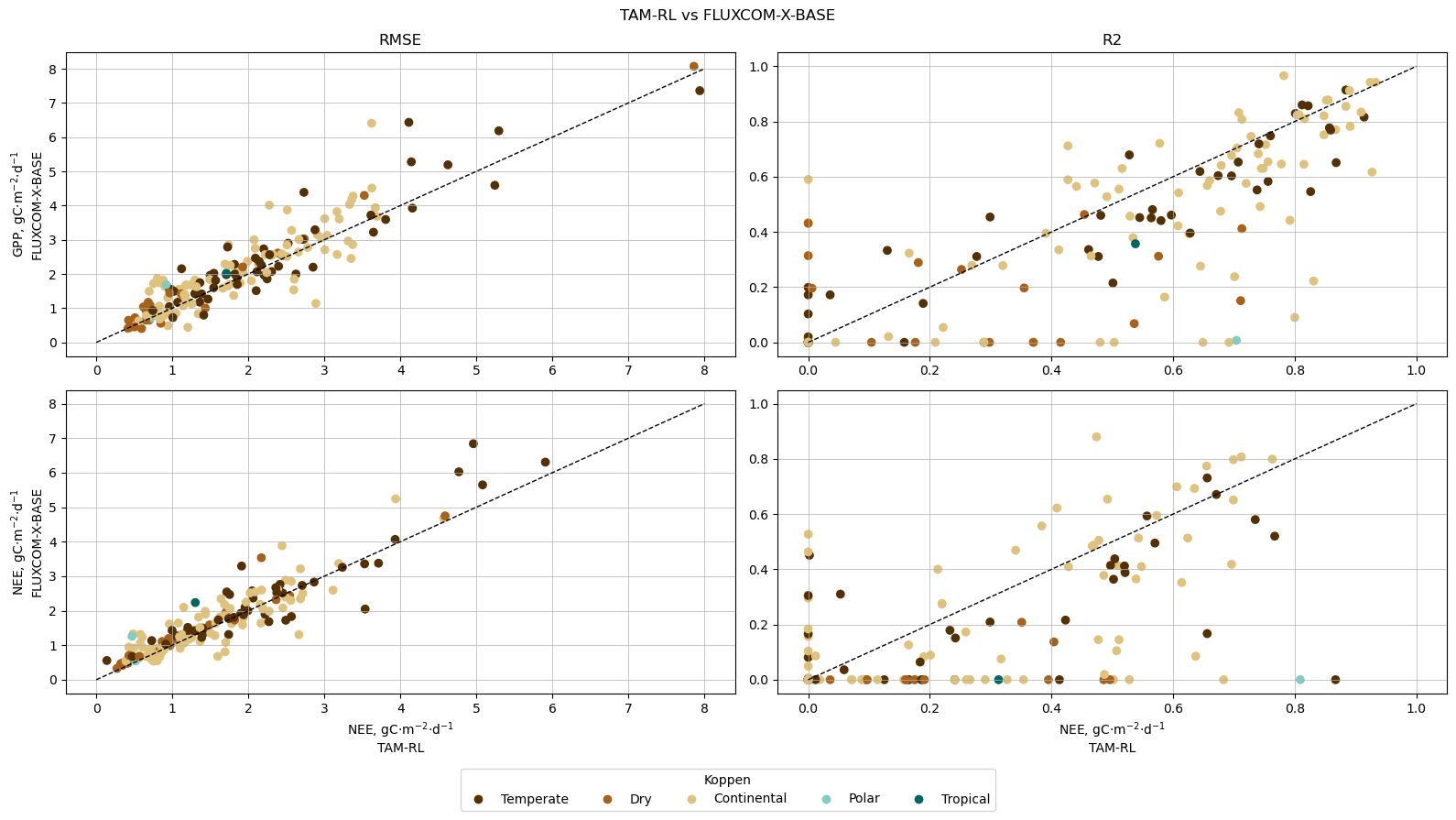}
    \caption{Scatter plots of RMSE and $R^2$ of TAM-RL and FLUXCOM-X-BASE computed for every site and colored by the Koppen climate type.}
    \label{fig:scatter}
\end{figure*}

Models were trained on the site list used in FLUXCOM-X-BASE and evaluated on the remaining 164 held-out sites. Fig.~\ref{fig:bars} shows RMSE distributions across IGBP ecoregions and K\"{o}ppen--Geiger climate zones, computed on highest-quality observations (NEE\_VUT\_USTAR50\_QC~$= 1$); error bars reflect within-category variance across sites. Across the five dominant climate types, CT-LSTM, TAMLSTM, and TAM-RL consistently outperform XGBoost and FLUXCOM-X-BASE, though performance degrades for certain IGBP classes such as water bodies (WAT).

Fig.~\ref{fig:heat} presents a relative RMSE heatmap computed as $(y_{fluxcom-x-base} - \hat{y}) / y_{fluxcom-x-base}$, offering a comparative rather than absolute view of improvement, with site and sample counts shown per IGBP type. Fig.~\ref{fig:scatter} provides per-site RMSE and $R^2$ comparisons between TAM-RL and FLUXCOM-X-BASE: points above the diagonal in the RMSE plot indicate improved accuracy, while points below the diagonal in the $R^2$ plot indicate higher explained variance.

\begin{table}[h]
\centering
\caption{\small Model performance comparison across targets. Metrics are averaged across all sites. NEE\_VUT\_USTAR50\_QC=1 and NEE\_VUT\_REF\_QC=1.}
\vspace{-0.35cm}
\label{tab:results}
\renewcommand{\arraystretch}{1.2}
\setlength{\tabcolsep}{6pt}

\begin{tabular}{l|cc|cc}
\hline
Model & \multicolumn{2}{c|}{\textbf{GPP}} & \multicolumn{2}{c}{\textbf{NEE}} \\
 & RMSE & $R^2$ & RMSE & $R^2$ \\
\hline
CT-LSTM        &  2.03  &  0.42  &  1.67  &  0.21  \\
TAMLSTM        &  2.04  &  0.40 & 1.63  &  0.21  \\
TAM-RL         &  \textbf{1.97}  &  \textbf{0.43} &  \textbf{1.62}  &  \textbf{0.23} \\
XGBoost        &  2.17  &  0.34  &  1.76  &  0.16  \\
FLUXCOM-X-BASE &  2.18  &  0.36  &  1.76  &  0.16  \\
\hline
\end{tabular}
\end{table}

\subsection{Summary}

Tab.~\ref{tab:results} aggregates average performance metrics across all sites for each target and model. TAM-RL consistently achieves the lowest RMSE and highest $R^2$ across both targets: compared to FLUXCOM-X-BASE, our framework reduces RMSE by 9.6\% for GPP and 8.0\% for NEE, while improving $R^2$ by 19.4\% and 43.8\%, respectively.

Several limitations remain. All models, including the XGBoost baseline, underperform on water bodies (WAT), suggesting the current feature set lacks variables capturing aquatic processes. A moderate performance drop is also observed for mixed forests (MF), deciduous broadleaf forests (DBF), and evergreen needleleaf forests (ENF), where RMSE differences relative to FLUXCOM-X-BASE range from 0.4--6.0\% depending on the target.

More broadly, our analysis reveals substantial error variability across IGBP and K\"{o}ppen--Geiger classes. While TAM-RL reduces this variance relative to baselines, spatial and climatic heterogeneity remain significant challenges for reliable global carbon flux upscaling.

\section{Conclusion}
\vspace{-0.1cm}
We introduced TAM-RL, a task-aware representation learning framework for zero-shot upscaling terrestrial carbon fluxes, trained with a knowledge-guided composite loss. Across all experiments, TAM-RL on average outperforms FLUXCOM-X-BASE and conventional baselines, achieving lower RMSE and higher $R^2$ for both GPP and NEE, while effectively capturing site-specific characteristics and temporal dynamics in the presence of sparse, heterogeneous EC data.

Challenges remain for underrepresented ecosystems, particularly water bodies and certain forest types, where performance lags behind, and substantial error variability persists across climate and ecoregion classes. Future work should focus on further developing methods for a better knowledge transfer between different locations as well as investigate variational or Bayesian extensions to better quantify and reduce predictive uncertainty in global carbon flux upscaling.

\section{Acknowledgment}
We acknowledge funding from NSF through the Learning the Earth with Artificial intelligence and Physics (LEAP) Science and Technology Center (STC) (Award \#2019625) and NSF Grant 2313174. Computational resources were provided by the Minnesota Supercomputing Institute. A. Renganathan acknowledges support from the University of Minnesota Doctoral Dissertation Fellowship (2025–26).
\bibliography{reference}

\end{document}